\documentclass[10pt,twocolumn,letterpaper]{article}

\usepackage{iccv}
\usepackage{times}
\usepackage{epsfig}
\usepackage{graphicx}
\usepackage{amsmath}
\usepackage{amssymb}
\usepackage{booktabs}
\usepackage{array}
\usepackage{tablefootnote}
\usepackage{array,multirow}

\newcommand{\bR}{\textbf{R}} %
\newcommand{\bD}{\textbf{D}} %
\newcommand{\bt}{\textbf{t}} %

\newcommand{\bp}{\textbf{p}} %
\newcommand{\bx}{\textbf{x}} %
 
\newcommand{\bK}{\textbf{K}} %
\newcommand\norm[1]{\left\lVert#1\right\rVert}

\newcommand{\Km}{\textbf{K}_{\alpha}} %
\newcommand{\Rm}{\textbf{R}_{\alpha}} %
\newcommand{\tm}{\textbf{t}_{\alpha}} %
\newcommand{\Cm}{\textbf{C}_{\alpha}} %
\newcommand{\Mm}{\textbf{M}_{\alpha}} %

\newcommand{\Dm}{\textbf{D}_{\theta}} %
\newcommand{\Fm}{\textbf{F}_{\delta}} %

\newcommand{\tp}[1]{#1^{\top}} %
\newcommand{\itp}[1]{#1^{-\top}} %
\newcommand{\cL}{\mathcal{L}} %

\newcommand{\dn}{\textit{DepthNet}} %
\newcommand{\cn}{\textit{CameraNet}} %
\newcommand{\fn}{\textit{FlowNet}} %
\newcommand{\gln}{\textbf{GLNet}}

\DeclareMathOperator*{\argmin}{arg\,min} %

\newcommand{\bF}{\textbf{F}} %

\newcommand{\vtext}[1]{\rotatebox[origin=b]{90}{#1}}

\usepackage[pagebackref=true,breaklinks=true,letterpaper=true,colorlinks,bookmarks=false]{hyperref}

\iccvfinalcopy %

\def\iccvPaperID{0} %

\ificcvfinal\fi
\begin{document}

\title{Self-supervised Learning with Geometric Constraints in Monocular Video\\
\emph{Connecting Flow, Depth, and Camera}}

\author{Yuhua Chen$^{1,2}$\hspace{10mm}Cordelia Schmid$^1$\hspace{10mm}Cristian Sminchisescu$^1$\\[2mm]
$^1$Google Research\hspace{10mm}
$^2$ETH Zurich\\[-1.5pt]
{\tt\small yuhua.chen@vision.ee.ethz.ch, \{cordelias, sminchisescu\}@google.com}
}

\maketitle

\begin{abstract}
We present \emph{\textbf{GLNet}}, a self-supervised framework for learning depth, optical flow, camera pose and intrinsic parameters from monocular video -- addressing the difficulty of acquiring realistic ground-truth for such tasks.
We propose three contributions: 1) we design new loss functions that capture multiple geometric constraints (\eg epipolar geometry) as well as an adaptive photometric loss that supports multiple moving objects, rigid and non-rigid, 2) we extend the model such that it predicts camera intrinsics, making it applicable to uncalibrated video, and 3) we propose several online refinement strategies that rely on the symmetry of our self-supervised loss in training and testing, in particular optimizing model parameters and/or the output of different tasks, thus leveraging their mutual interactions. The idea of jointly optimizing the system output, under all geometric and photometric constraints can be viewed as a dense generalization of classical bundle adjustment. We demonstrate the effectiveness of our method on KITTI and Cityscapes, where we outperform previous self-supervised approaches on multiple tasks. We also show good generalization for transfer learning in YouTube videos. 
\end{abstract}

\section{Introduction}
One of the fundamental computer vision problems is estimating the 3D geometry of dynamic scenes captured by a moving camera. This encompasses many visual tasks such as depth estimation, optical flow, odometry, \etc. Robust solutions would support a wide range of applications in autonomous driving, robotics, augmented reality, and scene interaction. 

The 3D visual reconstruction problem has been studied extensively. At one end, there are structure from motion systems that leverage sparse, hand-crafted geometric features (\eg SIFT~\cite{lowe1999object} or SURF~\cite{bay2006surf}), exact linear mathematical relations~\cite{hartley2003multiple} (\eg the epipolar constraint and the fundamental matrix for the two view geometry), as well as the non-linear refinement of the structure and motion outputs under geometric re-projection losses based on bundle adjustment~\cite{triggs1999bundle}. Impressive real-time processing pipelines of this type are now available, but they still offer sparse reconstructions, are affected by partial occlusion, and face difficulties in cases when the geometry of the scene or the motion is degenerate. 

At the other end, recent years have witnessed the development of deep learning techniques where structure and motion estimation is formulated in a supervised setting~\cite{eigen2014depth,kendall2015posenet}, as a pure prediction problem with little reference to exact geometric relations. This produces dense 3D estimates but comes with the inherent limitation of possible geometric inconsistency, the dependence on large training sets, domain sensitivity, and the difficulty of collecting good quality real-world ground-truth using special equipment like laser or stereo rigs with sensitive calibration~\cite{geiger2013vision}. Another option is to use synthetic data~\cite{mayer2016large}, however bridging the realism gap can be challenging. 

More recently,  several authors~\cite{godard2017unsupervised,zhou2017unsupervised} focused on designing self-supervised systems with training signals resulting from photometric consistency losses between pairs of images. These systems are effective and implicitly embed basic 3D geometry -- as points in the first view are back-projected in 3D, displaced by the camera motion and reprojected in the second view --, but still lack the fundamental geometric constraints that link the rigid and apparent motion in multiple views. 

In this work we introduce a self-supervised geometric learning framework, \textbf{GLNet}, which aims to integrate the advantages of modern deep-learning based self-supervised systems -- \emph{(a)} training without labeled data, \emph{(b)} offering dense reconstructions where prior knowledge can be automatically incorporated, and \emph{(c)} leveraging multiple inter-connected tasks, with the ones of classical structure from motion  -- \emph{(i)} explicitly representing exact mathematical relations (\eg the epipolar constraint) that always hold for rigid scenes, \emph{(ii)} being able to jointly refine all outputs, including depth, pose and camera intrinsics, under photometric and geometric constraints, as in bundle adjustment, and \emph{(iii)} breaking asymmetries between training and testing.

The proposed framework is extensively evaluated on KITTI and Cityscapes, where we achieve performance gains over the state-of-the-art. Moreover, our framework shows good performance in a transfer learning setting, and is able to robustly learn from uncalibrated video. %

\section{Related Work}
Understanding the geometry and dynamics of 3D scenes usually involves estimating multiple properties such as depth, camera motion and optical flow. \emph{Structure from motion} (SfM)~\cite{furukawa2010towards,hartley2003multiple,newcombe2011dtam,  schonberger2016structure, triggs1999bundle, wu2011visualsfm} or scene flow estimation~\cite{menze2015object,vedula1999three} are well-established methodologies with a solid record of fundamental and practical progress. Models for structure and motion estimation based on feature matching and geometric verification have produced impressive results~\cite{schonberger2016structure}, but their reconstructions are often sparse and prone to error in textureless or occluded areas.

To address the bottleneck of accurate feature matching, recent work has focused on deep-learning approaches for geometric reasoning. Several methods train networks based on ground-truth labels and have been successfully applied to many tasks, such as monocular depth prediction~\cite{eigen2015predicting,eigen2014depth,laina2016deeper,li2015depth}, optical flow~\cite{dosovitskiy2015flownet,ilg2017flownet,ranjan2017optical,sun2018pwc} and camera pose estimation~\cite{kendall2016modelling, kendall2017geometric,kendall2015posenet}. To leverage the multiple cues, the tasks can be tackled jointly by fusing boundary detection~\cite{ilg2018occlusions}, the estimation of surface normals~\cite{li2015depth}, semantic segmentation~\cite{chen2019learning,xu2018pad, zhang2018joint}, \etc.

To train with ground-truth labels, different authors rely either on specialized equipment (\eg LIDAR) for data collection~\cite{geiger2013vision, geiger2012we} which is expensive in practice, or on synthetic dataset creation~\cite{mayer2016large} which in many cases leads to domain shift~\cite{atapour2018real,chen2018domain,chen2018road}. To reduce the amount of ground-truth labels required for training, unsupervised approaches recently emerged. The core idea is to generate a differentiable warping between two images and use the photometric difference as proxy supervision. This has been applied to stereo images~\cite{garg2016unsupervised,godard2017unsupervised, zhan2018unsupervised} and optical flow~\cite{jason2016back, meister2018unflow, wulff2018temporal}.

In this work, we focus on learning from monocular video. Many recent methods go along similar lines. Zhou~\etal~\cite{zhou2017unsupervised} couple the learning of monocular depth and ego-motion. Vijayanarasimhan~\etal~\cite{vijayanarasimhan2017sfm} additionally learn rigid motion parameters of multiple objects. Subsequent methods further proposed improvements based on various techniques, such as an ICP alignment loss~\cite{mahjourian2018unsupervised}, supervision from SfM algorithms~\cite{klodt2018supervising}, optical flow~\cite{yin2018geonet, zou2018df}, edges~\cite{yang2018lego}, modeling multiple rigid motions informed by instance segmentation~\cite{casser2018depth}, motion segmentation~\cite{ranjan2019competitive}, minimum projection loss~\cite{godard2018digging}, \etc.

\begin{figure*}[h]
  \center
  \includegraphics[trim={0cm 3cm 0cm 0cm},clip,width=0.95\textwidth]{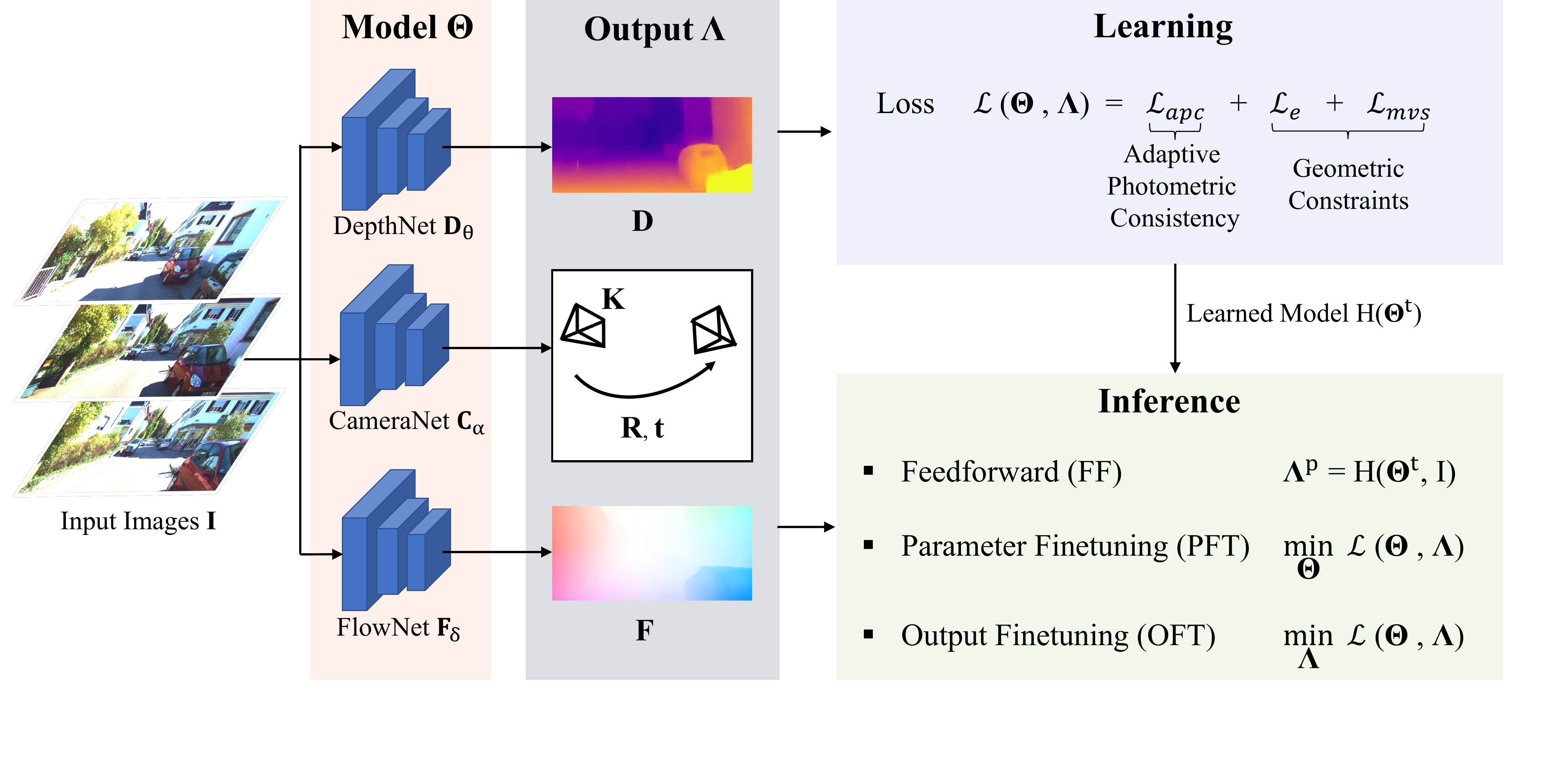}
  \caption{\textbf{Overview of the proposed \textbf{GLNet} framework.} Our model can take consecutive image frames as input, solves several tasks including depth, camera and flow estimation, and couples them via loss functions that capture the adaptive photometric and geometric constraints among outputs. An important feature of the proposed architecture is its training-testing symmetry as we learn a model in the training stage and during testing we further refine the parameters and the outputs based on the same optimization objective.} %
  \vspace{-4mm}
  \label{fig:overview}
\end{figure*}

\section{Methodology}

\paragraph{Overview} Our Geometric Learning Network (\textbf{GLNet}), for which an overview is presented in fig.~\ref{fig:overview}, solves the inter-related tasks of monocular depth prediction, optical flow, camera pose and intrinsics estimation, by relying on predictors for depth $\Dm$, camera $\Cm$ and flow $\Fm$ within a coupled optimization objective. $\Dm$ estimates the depth map $\bD$ from a single image. $\Cm$ predicts the camera pose $(\bR, \bt)$ between two adjacent frames, and also the intrinsic parameters $\bK$ whenever unknown. $\Fm$ operates on two image frames and estimates the optical flow. The joint output space of the tasks is denoted as $\boldsymbol{\Lambda} = \{\textbf{D}, \bR, \bt, \bK, \bF \}$, where the superset of parameters is $\boldsymbol{\Theta} = \{\boldsymbol{\theta}, \boldsymbol{\delta}, \boldsymbol{\alpha}\}$. The predictors $\{\Dm,\Cm,\Fm\}$ are implemented as three neural networks with trainable parameters $\boldsymbol{\Theta}$, referred to as \dn, \cn~ and \fn~ respectively. 

We formulate an optimization objective which consists of two parts: an adaptive photometric loss that captures the appearance similarity of the static and dynamic structures, and a geometric loss consisting of several components that couple the rigid and apparent motion. In training, the optimization objective is used as a proxy supervision signal to learn the parameters $\boldsymbol{\Theta}$ of the predictor. During inference, we can further refine the predictions $\boldsymbol{\Lambda}$ based on the same objective used in training, and we are able to focus on either refining the model parameters $\boldsymbol{\Theta}$ or the outputs $\boldsymbol{\Lambda}$, not unlike in geometric bundle adjustment. This is supported by our self-supervised objective and our explicit representation of classical geometric constraints.

\subsection{Geometric and Appearance Fundamentals}

 Consider a source image $I^s$ and a target image $I^t$, collected by a potentially moving camera with intrinsics $\Km$ and ego-motion $\Mm$. The 3D
 rigid transformation can be represented in homogeneous coordinates, in terms of a rotation matrix and a translation vector as

\begin{align}
  \label{eq:loss_avsl}
\Mm = \begin{bmatrix} 
\Rm & \tm\\
0 & 1 \\
\end{bmatrix}
\end{align}

For camera intrinsics, we fix the optical center at the image center, ignore camera skew and radial distortion, parameterize the focal length along the two optical axes as $f^x$ and $f^y$, and work with images of resolution $h \times w$. The intrinsics camera matrix writes as
\begin{align}
  \label{eq:loss_avsl}
\Km = \begin{bmatrix} 
f^x_{\alpha} & 0 & w/2 \\
0 & f^y_{\alpha} & h/2  \\
0 & 0 & 1 \\
\end{bmatrix}
\end{align}

Given a pixel $\bp$ in the source image $I^{s}$, the corresponding 3D scene position in the source camera coordinate system in homogeneous coordinates, can be back-projected as
\begin{align}
  \label{eq:bx}
        \bx = \begin{bmatrix} \Dm(\bp) \Km^{-1} \bp \\ 1 \end{bmatrix} 
\end{align}

Assuming we displace $\bx$ rigidly, the corresponding coordinates (projection) $\bp^{\prime}$ in the target image $I^{t}$, are

\begin{align}
  \label{eq:p_rigid}
{\bp^{\prime}} = \big[\Km | 0 \big] \Mm \begin{bmatrix} \Dm(\bp) \Km^{-1} \bp \\ 1 \end{bmatrix}
\end{align}

In this case, the displacement field $\bp^{\prime}-\bp$ represents the 2D projection of the 3D \emph{scene flow}, \ie the real image motion induced by the true underlying 3D rigid displacement. 

For pixels that cannot be explained by a rigid transformation, another key quantity is the measured, apparent motion or \emph{optical flow} $\Fm$ that provides a dense correspondence field between the source and the target images, respectively
\begin{align}
\label{eq:p_dynamic}
\bp^{\prime} = \bp + \Fm (\bp)
\end{align}
For simplicity, and in a slight abuse of notation, we also use $\bp$ as the 2D (non-homogeneous) image coordinates in the above equation. 

\subsection{Optimization Objectives}

In this section we describe the loss we use for the self-supervised learning of depth, flow, and camera matrix. These tasks are interconnected through photometric and geometric constraints.  
\paragraph{Adaptive Photometric Loss}
The view synthesis loss is widely used in self-supervised learning~\cite{zhou2017unsupervised}. 
The loss measures the photometric difference between the synthesized and the actual image, where synthesis is obtained by a 3D reconstruction in the first frame with backprojected pixel intensities, followed by rigid displacement and perspective projection in the second frame. However, this displacement would only be valid for scene structures consistent with the ego-motion, or moving according to a global rigid displacement. For secondary or non-rigidly moving objects, an adaptive approach is necessary, and we pursue one here.

Given source and target images $I^{s}$ and $I^{t}$, their pixels belong either to regions explained by ego-motion (or global scene motion), or by secondary dynamic objects.  %
For scene structures not explained by the global rigid motion, one can rely on the more flexible optical flow. The main intuition of our adaptive photometric loss is to channel parameter updates towards only those configurations that best explain the displacement, be it produced by either ego-motion or secondary displacements. The term can thus be represented as the minimum photometric error between the two displacements (\ie optical flow and rigid motion), as follows
\begin{align}
  \label{eq:loss_avsl}
    \cL_{apc}  = \min\Bigg\{&S\Big(I^{s}(\bp), I^{t}([\Km | 0 \big] \Mm \begin{bmatrix} \Dm(\bp) \Km^{-1} \bp \\ 1 \end{bmatrix} )\Big),   \nonumber \\  
    &S\Big(I^{s}(\bp), I^{t}(\bp + \Fm (\bp))\Big)\Bigg\}  
   \end{align}
   
\noindent The term is a per-pixel minimum between two components: the first part represents rigid displacement as in \eqref{eq:p_rigid}, and the second part represents the displacement produced by optical flow, \cf \eqref{eq:p_dynamic}. $S$ is a similarity function between two pixels. As common practice~\cite{yin2018geonet,zou2018df}, we use a weighted sum of structural similarities \emph{($SSIM$)}~\cite{wang2004image} and L1 components, with trade-off parameter $r (= 0.85)$, as follows
\begin{align}
  \label{eq:sim}
    S(a,b) = r\frac{{1-SSIM(a,b)}}{2}+(1-r)\|a-b\|_1
\end{align}

\vspace{-5mm}
\paragraph{Multi-view 3D Structure Consistency} 
Predicting depth independently for each view usually results in inconsistency among the same scene structure in different (\eg consecutive) views. To enforce structural consistency in 3D space, we design a loss component to penalize 3D structural deviations in multiple views.  

Given a pixel $\bp$ in the source image, and its corresponding pixel in the target image $\bp^{\prime}$ given by \eqref{eq:p_rigid}, their predicted 3D coordinates can be obtained by back-projection, \cf \eqref{eq:bx}. To obtain consistent structure, we penalize deviations between estimates of the same 3D point, once transformed to the common coordinate system of the target camera
\begin{align}
  \label{eq:loss_3D}
    \cL_{mvs} = & \| \bx^{\prime} - \bx  \|_1 =\\
            = & \norm{ \begin{bmatrix} \Dm(\bp^{\prime}) \Km^{-1} \bp^{\prime} \\ 1 \end{bmatrix} - \Mm  \begin{bmatrix} \Dm(\bp) \Km^{-1} \bp \\ 1 \end{bmatrix}   }_1  \nonumber 
\end{align}
The loss represents the 3D discrepancy of predictions from two views, and the gradients are used to update \cn ~$\{\Km, \Rm, \tm\}$ and \dn ~$\Dm$. Notice that such a loss can immediately generalize to multiple views.

\vspace{-3mm}
\paragraph{Epipolar Constraint Loss for Optical Flow}
The epipolar constraint is widely used in classical geometric methods~\cite{hartley2003multiple}, in order to compute a closed-form solution to initialize non-linear bundle adjustment procedures.
Notably absent from existing deep learning-based structure and motion prediction systems, the epipolar constraint is an algebraic relationship that couples 3D scene projections, \eg $\bp$ and $\bp^{\prime}$ in two views, via a fundamental or essential matrix, that embeds the geometry of the camera displacement $\textbf{M}$ and its intrinsic parameters $\bK$. An alternative view of the epipolar constraint, that we leverage in our model, is as a verification equation for putative correspondences, as provided, \eg by optical flow, \cf \eqref{eq:p_dynamic}.

Optical flow has been traditionally formulated as an optimization of a matching and smoothness loss, or more recently as per-pixel regression problem within deep-learning methods \cite{dosovitskiy2015flownet, ilg2017flownet}. However, results computed that way may be inconsistent with any epipolar geometry. 
To endow the learning process with geometric awareness, we incorporate the epipolar constraint as a penalty over dense correspondences computed by optical flow.  The resulting epipolar constraint loss writes 
\begin{align}
  \label{eq:loss_epipolar}
\cL_{e} = \tp{\bp} \itp{\Km} \Rm \big[ \tm \big]_{\times} \Km^{-1} (\bp + \Fm(\bp))
\end{align}

\noindent where $\left[\bt\right]_{\times}$ is the skew-symmetric matrix corresponding to the translation vector $\bt$. The loss enforces a global epipolar constraint over the dense correspondences from optical flow, and the associated gradients are used to update  \cn ~$\{\Km, \Rm, \tm\}$ and \fn ~$\Fm$.

\vspace{-4mm}
\paragraph{Regularization}
We also rely on standard regularization terms $\cL_{r}$ for our learning framework. Specifically, we use separate spatial smoothness terms over depth $\Dm$, and optical flow $\Fm$ respectively, as well as forward-backward flow consistency constraints. 

\vspace{-3mm}
\paragraph{Total Loss} 
The total loss is a weighted sum over losses previously introduced, each densely summing over image pixels, with weighting dropped for simplicity
\begin{align}
\label{eq:loss_overall}
 \cL(\boldsymbol{\Lambda}, \boldsymbol{\Theta}) = \cL_{apc} + \cL_{mvs} + \cL_{e} + \cL_{r} 
\end{align}

The model is trained jointly in an end-to-end manner.

\subsection{Online Optimization}
During inference, most existing approaches produce results by running trained models feed-forward. This can be problematic as the structural constraints between multiple outputs, or between inputs and outputs, \eg image intensity, flow, camera motion and depth, are no longer preserved.

Our solution is to operate with -- and optimize over -- similar losses during both learning and inference, \cf \eqref{eq:loss_overall}. This is possible as our objective is self-supervised, hence it eliminates the asymmetry between training and testing. Therefore an online refinement process, either for model parameters, or for model outputs, is possible, and can leverage task dependencies for optimal performance and seamless model adaption to new environments. 

Formally, we optimize our model on the training set in order to obtain initial estimates $\boldsymbol{\Theta}^t$. Giving a test image pair from a monocular video, initial predictions are computed, feed-forward, in the standard way, as $\boldsymbol{\Lambda}^p = \{{\textbf D}^p,\bF^p,\bR^p,\bt^p, \bK^p\}$. The (self-supervised) objective in \eqref{eq:loss_overall} can be computed, as no ground-truth label is needed. To prevent overfitting to the optimization objective, based on a single image pair, we further enforce a regularizer on the output space $ \| \boldsymbol{\Lambda} - \boldsymbol{\Lambda}^p\|$ in order to penalize large deviation from the the original prediction of the learned model. The full optimization objective writes as
\begin{align}
\label{eq:optimization}
\{\boldsymbol{\Lambda}^{*},\boldsymbol{\Theta}^{*}\} = \argmin_{\boldsymbol{\Theta}, \boldsymbol{\Lambda}} \cL(\boldsymbol{\Theta}, \boldsymbol{\Lambda}) + \| \boldsymbol{\Lambda} - \boldsymbol{\Lambda}^p\|  + \| \boldsymbol{\Theta} - \boldsymbol{\Theta}^t\|
\end{align}

To update the output, we design two strategies, which can be used together or independently. In \textbf{Parameter Finetuning (PFT)} we update the parameters of the predictor $\boldsymbol{\Theta}$ based on the loss. In \textbf{Output Finetuning (OFT)} we optimize the output directly without recomputing the network but with the trained prediction as prior. This is implemented by considering the outputs $\boldsymbol{\Lambda}$ as free variables initialized as the predictions of the trained neural network, and using the gradient of the loss to refine them. Besides initialization, the computational graph associated to the deep model is no longer used, and only one forward pass through the deep model is needed. Moreover, the number of output optimization variables is typically much lower than the number of network parameters. Thus OFT offers the benefit of increased speed -- in practice one order of magnitude. 

\begin{table*}[t]
\begin{center}
\resizebox{\textwidth}{!}{%
\begin{tabular}{c|c|c|c|c|c|c|c|c|c}
\toprule
& Method & Calib. & Abs Rel & Sq Rel & RMSE & RMSE log & $\delta<1.25$ & $\delta<1.25^2$ & $\delta<1.25^3$\\
& & & $\downarrow$ & $\downarrow$ & $\downarrow$ & $\downarrow$ & $\uparrow$ & $\uparrow$  & $\uparrow$ \\
\hline
\hline
\multirow{11}{*}{\vtext{Trained on \textbf{KITTI}}}
& { Zhou \etal* ~\cite{zhou2017unsupervised}} & \checkmark &0.183 & 1.595 & 6.709 & 0.270 & 0.734 & 0.902 & 0.959 \\
& { Mahjourian \etal~\cite{mahjourian2018unsupervised}}& \checkmark& 0.163 & 1.240 & 6.220 & 0.250 & 0.762 & 0.916 & 0.968 \\
& { GeoNet~\cite{yin2018geonet} }& \checkmark&{0.155} & 1.296 & 5.857 & 0.233 & {0.793} & {0.931} & {0.973} \\
& { Wang \etal \cite{wang2018learning}} & \checkmark& 0.151 & 1.257 & 5.583 & 0.228 & 0.810 & 0.936 & 0.974 \\
& { LEGO \cite{yang2018lego}} & \checkmark& 0.162 & 1.352 & 6.276 & 0.252 & - & - & -\\
& { DF-Net \cite{zou2018df} }& \checkmark& 0.150  &1.124&  5.507& 0.223& 0.806& 0.933& 0.973 \\
& {Casser \etal\cite{casser2018depth}} & \checkmark& 0.109 & 0.825 & 4.750 & 0.187 & 0.874 & 0.957 & \textbf{0.982} \\

\cline{2-10}
& \textit{Baseline} & \checkmark& 0.156 & 1.450 & 5.913 & 0.228 & 0.861 & 0.931 & 0.955\\

& \gln (-ref.)& \checkmark  & 0.135  & 1.070 & 5.230 & 0.210 & 0.841 & 0.948 & 0.980\\

&{\gln}        &  $\times$  &0.100  &0.811   &4.806   &0.189    &0.875    &\textbf{0.958}    &\textbf{0.982} \\
&{\gln} & \checkmark& \textbf{0.099} & \textbf{0.796} & \textbf{4.743} & \textbf{0.186} & \textbf{0.884} & {0.955} & {0.979} \\

\hline\hline
\multirow{5}{*}{\vtext{Trained on \textbf{CS}}}
&{GeoNet}~\cite{yin2018geonet} & \checkmark  & 0.210 & 1.723 & 6.595 & 0.281 & 0.681 & 0.891 & 0.960 \\

&{Casser \etal\cite{casser2018depth}} & \checkmark & 0.153 & 1.109 & 5.557 & 0.227 & 0.796 & 0.934 & 0.975 \\

\cline{2-10}
& \textit{Baseline} & \checkmark & 0.206 & 1.611 & 6.609 & 0.281 & 0.682 & 0.895 & 0.959 \\

& \gln & $\times$ & 0.144 & 1.492 & 5.473 & 0.219 & 0.831 & 0.932 & 0.967 \\

& \gln& \checkmark & \textbf{0.129} & \textbf{1.044} & \textbf{5.361} & \textbf{0.212} & \textbf{0.843} & \textbf{0.938} & \textbf{0.976}\\ 

\bottomrule
\end{tabular}}
\caption{\textbf{Results of depth estimation on the KITTI Eigen Split.} We report models trained on KITTI and Cityscapes. The best result in each setting is marked in bold. '\textbf{GLNet}(-ref.)' denotes the results without online refinement. *The result of Zhou \etal~\cite{zhou2017unsupervised} is based on their updated Github version.}
\label{tab:depth}
\end{center}
\end{table*}

\begin{figure*}[t]
\small
\centering
\input{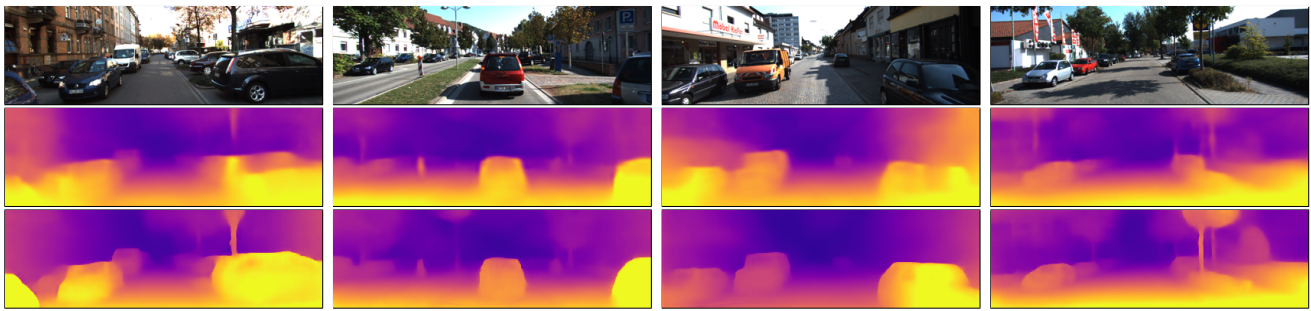}
  \caption{\textbf{Qualitative results of depth estimation.} \textit{Top Row:} Input images, \textit{Middle Row:} Baseline results, \textit{Bottom Row:} \textbf{GLNet} results. The proposed framework offers sharper predictions.}
\label{fig:depth_qualitative}
\end{figure*}

\subsection{Network Architecture}
The focus of this work is on different loss functions and fine-tuning options, hence our network design mostly aligns with existing self-supervised learning components for geometric processing. Here we briefly cover the architectures used in this work.

\vspace{-4mm}
\paragraph{\dn} maps an input image to a per-pixel depth map. It is based on a fully convolutional encoder-decoder structure. The encoder is based on ResNet18~\cite{he2016deep}. The decoder relies on DispNet~\cite{mayer2016large}, and consists of several deconvolutional layers. Skip connections are used to provide spatial context, and depth is predicted at four different scales.  

\vspace{-4mm}
\paragraph{\cn} takes two adjacent image frames as input, and regresses the 6DOF camera pose, represented as a translation vector and a relative rotation matrix parameterized in terms of three Euler angles. When learning from uncalibrated video, the network also predicts camera intrinsics.  We use the model of \cite{zhou2017unsupervised} which is a small network with 8 convolutional layers. A global pooling on the last layer is used for the final prediction.

\vspace{-4mm}
\paragraph{\fn} predicts the optical flow $\bF$ between two adjacent image frames. We use the same architecture described in \cite{yin2018geonet}, which is an encoder-decoder with a ResNet backbone. 

\vspace{1mm}
Notice that our framework is agnostic to the particular choice of each component network, and other options are possible. Hence we can benefit from higher performing networks for individual tasks. 
\vspace{-1mm}

\section{Experiments}

In this section, we validate \textbf{GLNet} through extensive experiments on depth, optical flow, and camera pose estimation. We first introduce the datasets and the parameter settings used in experiments. 

\vspace{-3mm}
\paragraph{Dataset} Our experiments are mainly conducted on KITTI~\cite{geiger2013vision, geiger2012we} and Cityscapes~\cite{cordts2016cityscapes}. KITTI is a widely used dataset for benchmarking geometric understanding tasks such as depth estimation, odometry, and optical flow. Images are captured using cameras mounted on cars. We evaluate using the ground-truth labels provided with the official KITTI dataset. We additionally train the framework on Cityscapes~\cite{cordts2016cityscapes}, and study how well the proposed models transfer across datasets. Similar with KITTI, Cityscapes mainly contains data collected by cars driving in European cities. 

\vspace{-3mm}
\paragraph{Parameter Settings} In training, we use Adam~\cite{kingma2014adam} with $\beta_1=0.9$ and $\beta_2=0.999$. The initial learning rate is set to $2\times 10^{-4}$ and batch size to 4. The images are resized to $128\times 416$ resolution. Each training sample is a snippet consisting of three consecutive frames. Additionally, random resizing, cropping, flipping and color jittering are used during training, as common practice, for data augmentation. The backbone network is initialized with ImageNet weights, and we optimize the network for maximum 30 epochs although convergence usually occurs earlier.

For online refinement, we initialize the models with weights learned in the training stage. We use a batch size of $1$. The batch consists of the test image and its two adjacent frames. Online refinement is performed for $50$ iterations on one test sample with the same hyper-parameter introduced before. No data augmentation is used in the inference phase.

In all experiments, the model with only standard (non-adaptive) photometric loss is used as the baseline.

\subsection{Depth Estimation}
We begin with the evaluation of depth estimation. As common in protocols previously used~\cite{zhou2017unsupervised}, we report results of depth estimation using the Eigen~\cite{eigen2014depth} split of the raw KITTI dataset \cite{geiger2013vision}, which consists of $697$ test images. Frames that are similar to the test scenes are removed from the training set. We compare the performance of the proposed framework with the baseline, as well as recent state-of-the-art works in the same setting~\cite{ casser2018depth,mahjourian2018unsupervised, wang2018learning, yang2018lego, yin2018geonet, zhou2017unsupervised, zou2018df}. 

\vspace{-3mm}
\paragraph{Main Results} As shown in table \ref{tab:depth}, our method achieves significant gains over the baseline, as well as the other competing methods, when using ground-truth camera intrinsics (as typical). Qualitative results are shown in fig.~\ref{fig:depth_qualitative}, where we observe a clear improvement in the visual quality of the depth map. 

We notice that the online refinement is not used in other methods except \cite{casser2018depth}. Thus to facilitate the comparison, we also report the results of \textbf{GLNet} without online refinement, denoted as \textbf{GLNet}(-ref.). Still, this yields better performance than the competing methods without refinement~\cite{mahjourian2018unsupervised,wang2018learning,yin2018geonet,zhou2017unsupervised,zou2018df}, demonstrating the effectiveness of our proposed loss functions. 

In the uncalibrated scenario, where no ground-truth camera intrinsics are given, we use our \cn~ to predict them from input images. The resulting performance matches the calibrated setting. This is a sanity check but not entirely surprising as the camera used to collect the test set has a similar setup with the ones used in training. 

The framework is also tested in a transfer learning setting. We train the models self-supervised on Cityscapes, then apply them on KITTI. Again, we observe performance gains when using the proposed components, and overall better results compared to competing methods. This shows the ability of the proposed framework to generalize to new environments. It also demonstrates that geometric constraints are powerful in closing domain gaps, as such mathematical relations are always valid. Interestingly, our method still displays competitive performance when no ground-truth camera calibration is given, even though the camera intrinsics of Cityscapes are different from those of KITTI.  

\begin{figure*}[t]
\small
\centering
\input{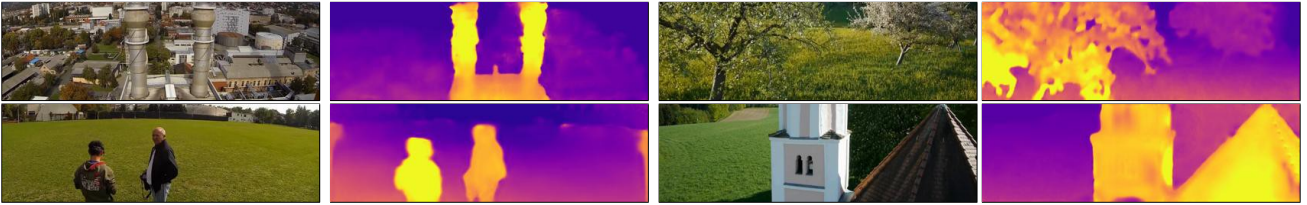}
  \caption{\textbf{Qualitative results on YouTube videos, where camera intrinsics are unknown.} \textit{Left:} Input images, \textit{Right:} Depth estimation results of \textbf{GLNet}. The proposed method can robustly estimate depth from uncalibrated video.}
\label{fig:youtube_qualitative}
\end{figure*}

\vspace{-3mm}
\paragraph{Generalization to Videos in the Wild}
To further verify the generalization ability of the proposed method, we test \textbf{GLNet} on a set of videos collected from YouTube, where the camera intrinsic parameters are unknown. We present qualitative results of \textbf{GLNet} in fig.~\ref{fig:youtube_qualitative}. As can be seen, the proposed method is able to estimate depth from uncalibrated video frames across a variety of objects, structures and scenes. This supports the claim that our method can generalize to uncalibrated videos.

\vspace{-3mm}
\paragraph{Ablation Study on Losses}
To analyze the individual impact of each loss component, we provide an ablation study over different combinations of losses. As shown in table~\ref{tab:depth_ablation}, our adaptive photometric loss achieves a healthy improvement over the standard photometric loss. 
The performance is further improved by using the geometric losses, especially when coupled with online refinement. Out of the three losses proposed, the multi-view structure consistency seems most effective for depth estimation, which is understandable as it is directly linked to 3D and integrates information over multiple views. 

\begin{table}[t]
\begin{center}
\begin{tabular}{c c c c | c }
\toprule
$\cL_{apc}$ & $\cL_{mvs}$ & $\cL_{e}$ & refinement & Abs Rel \\
\hline
 & & & & 0.156     \\
\checkmark & & & & 0.144         \\
\checkmark & \checkmark & & & 0.138         \\
\checkmark & \checkmark & \checkmark  &  & 0.135         \\
\hline
 & & &  \checkmark & 0.137     \\
\checkmark & & &  \checkmark & 0.130         \\
\checkmark & \checkmark & &  \checkmark & 0.103         \\
\checkmark & \checkmark  & \checkmark & \checkmark  & 0.099         \\
\bottomrule
\end{tabular}
\vspace{3mm}
\caption{\textbf{Ablation study on losses.} We evaluate ablated version of the proposed method on the Eigen split. Notice the improvements offered by the various components.}
\label{tab:depth_ablation}
\vspace{-3mm}
\end{center}
\end{table}

\begin{figure}[h]
  \center
\resizebox{0.95\linewidth}{!}{
    \includegraphics[]{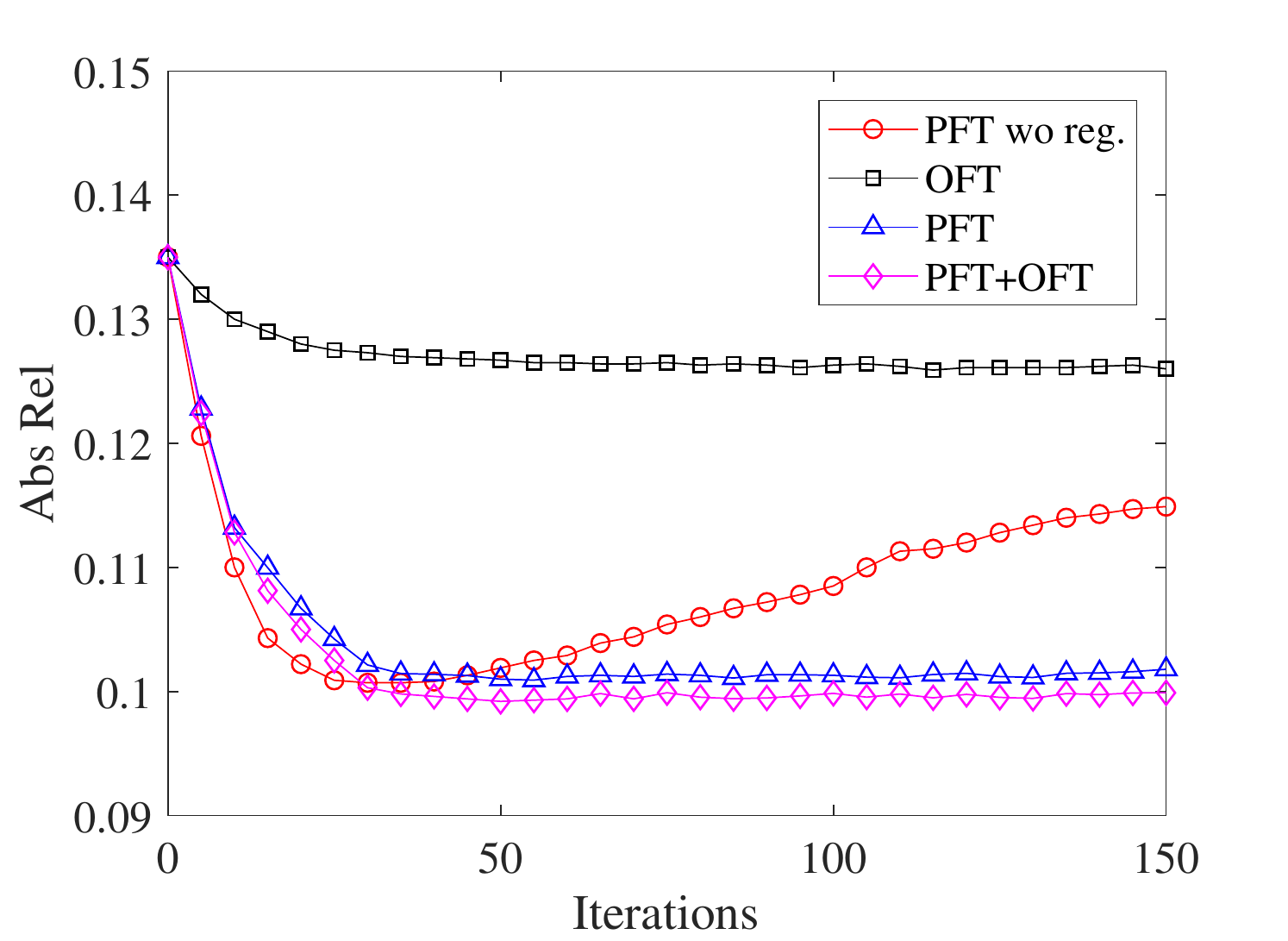}}
  \caption{\textbf{Study of refinement strategies.} We report the Abs Rel evaluation metric as a function of iteration. Our method achieves the best result and does not overfit to the test sample as refinement proceeds.}
  \label{fig:exp_iters}
\end{figure}

\vspace{-3mm}
\paragraph{Online Refinement}
We also conduct an ablation study over refinement strategies. We use the normal parameter finetuning without regularizer as baseline, as in \cite{casser2018depth}. We compare the baseline with the two proposed refinement strategies, \ie OFT and PFT. As shown in fig.~\ref{fig:exp_iters}, using standard finetuning we achieve some performance gain, however the downside is rapid overfitting to the test sample. The regularizer can effectively prevent the model from overfitting, for both PFT and OFT. The improvement from OFT is comparable to PFT. However, it is much faster compared to the other options as OFT only updates output variables, whose dimensionality is, for this problem, in the number of pixels $\sim 10k$. PFT needs to update the parameter of the neural network, whose size is around $\sim 1M$. At runtime, OFT takes about 2 seconds for 50 iterations, whereas PFT typically runs for around 40 seconds for the same number of iterations, which is \emph{one order of magnitude slower}. Combining PFT and OFT brings some improvement over PFT alone, and achieves the best overall performance.

\begin{figure*}[t]
\small
\centering
\input{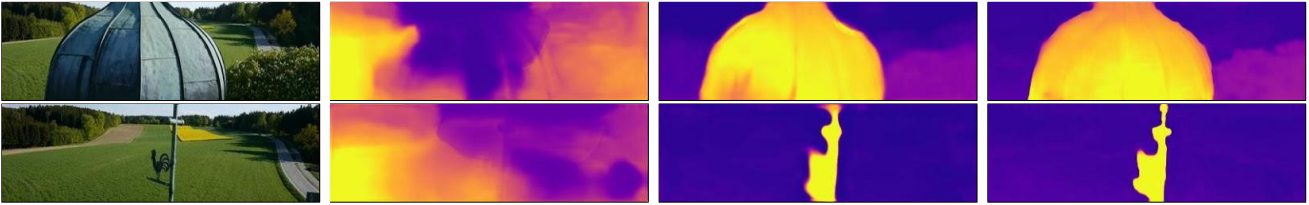}
  \caption{\textbf{Results of online refinement on YouTube videos.}  \textit{From left to right:} input images, feed-forward (FF) results, OFT results, PFT results. The results of online refinement, OFT, are noticeably better than the feed-forward model -- comparable to PFT but much faster.}
\label{fig:youtube_refinement}
\end{figure*}

We also note online refinement is especially effective in the transfer learning setting. As illustrated in fig.~\ref{fig:youtube_refinement}, the model trained on KITTI produces erroneous depth prediction, when running the standard feed-forward component in a new environment. The prediction results can be significantly improved by both PFT and OFT. PFT delivers results that appear slightly sharper, visually, than OFT, which is consistent with the quantitative results. 

\begin{figure*}[t]
\small
\centering
\input{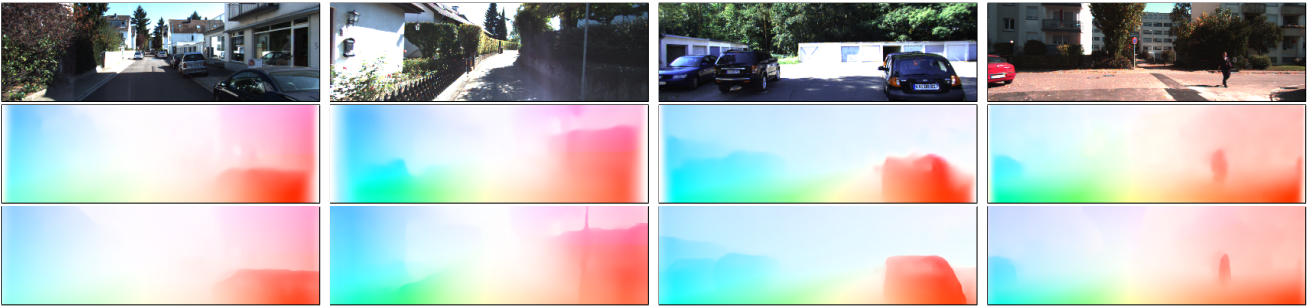}
  \caption{\textbf{Qualitative results of flow estimation.} \textit{Top Row:} Input images, \textit{Middle Row:} without geometric constraints, \textit{Bottom Row:} with geometric constraints. The geometric constraints significantly improve the quality of the predicted rigid component flow.}
\label{fig:flow_qualitative}
\end{figure*}

\subsection{Optical Flow}
To benchmark optical flow, we use the KITTI 2015 stereo/flow training set \cite{menze2015object} which has 200 training images. Similarly with previous work~\cite{yin2018geonet, zou2018df}, we use the training set for evaluation, as the proposed framework is self-supervised. We report the performance using the average end-point error (EPE) over non-occluded regions (Noc) and overall regions (All). 

As shown in table~\ref{tab:flow}, the proposed adaptive loss and multi-view structure consistency loss don't show much advantage over the baseline, possibly due to their indirect influence on the flow prediction. However, the epipolar constraint loss achieves a considerable performance gain over the baseline. Results can be further improved by online refinement.
Qualitative results are provided in fig.~\ref{fig:flow_qualitative}, where we observe that geometric constraints substantially improve the flow quality for rigidly moving scene regions. In an uncalibrated setting, the performance is very close to the calibrated one.  This is understandable, as the optical flow prediction does not rely on precise intrinsics but can benefit from geometric epipolar corrections which are however calibration-sensitive.

\begin{table}
\begin{center}
\setlength{\tabcolsep}{5.0pt}
\begin{tabular}{c | c c}
\toprule
Method & Noc & All \\
\hline
FlowNetS~\cite{dosovitskiy2015flownet} & 8.12 & 14.19 \\
FlowNet2~\cite{ilg2017flownet} &  4.93 & 10.06 \\
\hline
GeoNet~\cite{yin2018geonet} &  8.05 & 10.81  \\
DF-Net~\cite{zou2018df} &  - & 8.98 \\
\hline
\textit{Baseline} &  6.80 & 12.28 \\
$\cL_{apc}$ &  6.78 & 12.26 \\
$\cL_{apc} + \cL_{mvs}$ &  6.77 & 12.20 \\
$\cL_{apc} +  \cL_{mvs} + \cL_{e}$ &  5.40 & 8.95 \\
\gln &  \textbf{4.86} & \textbf{8.35} \\
{\bf GLNet} (uncalibrated) &  4.90 & 8.41 \\
\bottomrule
\end{tabular}
\end{center}

\caption{\textbf{Evaluation of optical flow.} We report average end-point error (EPE) on the KITTI 2015 flow training set over non-occluded regions (Noc) and overall regions (All). The first two methods are supervised, and other methods are trained on KITTI unsupervisedly. The best result is marked in bold.}
\vspace{-0mm}

\label{tab:flow}
\end{table}

\subsection{Pose Estimation Results}
We also evaluate the performance of our \textbf{GLNet} on the official KITTI visual odometry benchmark. As in the standard setting, we use the sequences 00-08 for training, and sequences 09-10 for testing. The pose estimation results are summarized in table~\ref{tab:pose}, showing improvement over existing methods, as well as our baseline. 

\begin{table}
\vspace{2mm}
\small
\begin{center}
\setlength{\tabcolsep}{5.0pt}
\begin{tabular}{c | c | c}
\toprule
Method & Seq.09 & Seq.10 \\
\hline
ORB-SLAM~(full) & $0.014\pm 0.008$ & $0.012\pm 0.011$\\
ORB-SLAM~(short) & $0.064\pm 0.141$ & $0.064\pm 0.130$\\
Zhou~\etal~\cite{zhou2017unsupervised} & $0.016\pm 0.009$ & $0.013\pm 0.009$\\
Mahjourian~\etal~\cite{mahjourian2018unsupervised} & $0.013\pm 0.010$ & $0.012\pm 0.011$\\
GeoNet~\cite{yin2018geonet} &0.012 $\pm$ 0.007& 0.012 $\pm$ 0.009\\
DF-Net~\cite{zou2018df} &0.017 $\pm$ 0.007& 0.015 $\pm$ 0.009\\
Casser \etal\cite{casser2018depth} &\textbf{0.011 $\pm$ 0.006}& \textbf{0.011 $\pm$ 0.010}\\
\hline
\textit{Baseline} &0.013 $\pm$ 0.007& 0.012 $\pm$ 0.010\\
\gln &\textbf{0.011 $\pm$ 0.006}& \textbf{0.011 $\pm$ 0.009}\\
\bottomrule
\end{tabular}
\end{center}

\vspace{1.5mm}
\caption{\textbf{Evaluation on camera pose estimation.} Absolute trajectory error (ATE) on KITTI odometry dataset is reported as the evaluation metrics. Best result is shown in bold.}
\vspace{-6mm}
\label{tab:pose}
\end{table}

\section{Conclusions}

We have presented \textbf{GLNet}, a geometrically-inspired learning framework to jointly learn depth, flow, camera pose and intrinsic parameters from monocular video. The model is self-supervised and combines novel photometric and geometric loss functions, some based on fundamental relations like the epipolar constraint -- and relying on continuously estimated correspondences from optical flow--, others on structure consistency over time. We've also introduced new parameter and output finetuning methods that generalize bundle adjustment, break the existing asymmetries between training and testing, and offer online adaption speed-ups of up to one order of magnitude. Given its ability to predict camera intrinsics, the model can be applied to uncalibrated video, and
exhibits consistent performance across different training and testing domains. This supports the conclusion that geometric constraints represent a valuable regularizer in a transfer learning setting. 

{\small
\bibliographystyle{ieee_fullname}
\bibliography{main_bib}
}

\end{document}